# Data-Driven Automated Identification of Optimal Feature-Representative Images in Infrared Thermography Using Statistical and Morphological Metrics.

Harutyun Yagdjian[a], Martin Gurka[a]

[a]*Leibniz-Institut für Verbundwerkstoffe GmbH, Erwin-Schrödinger-Straße 58, 67663 Kaiserslautern*

**Abstract**

Infrared thermography (IRT) is a widely used non-destructive testing technique for detecting structural features, such as subsurface. Most IRT data post-processing methodologies result in the generation of image sequences in which defect visibility varies strongly across time, frequency, or coefficient/index domains, making the selection of defect-representative images a non-trivial and critical task. Conventional evaluation metrics such as the signal-to-noise ratio (SNR) or the Tanimoto criterion often rely on prior knowledge of defect location or defect free reference regions, which limits their applicability for automated and unsupervised analysis. In this work, a data-driven methodology is proposed to identify images within IRT datasets that are most likely to contain and represent features, especially anomalies and defects, without requiring any prior information about their spatial position. The investigation focuses on three complementary metrics. Firstly the Homogeneity Index of Mixture (*HI*), which quantifies statistical heterogeneity through deviations of local intensity distributions from a global reference distribution. The second metric is a Representative Elementary Area (*REA*) derived from a Minkowski-functional-based adaptation of the Representative Elementary Volume (*REV*) concept to two-dimensional images. Building upon these approaches, a third metric is introduced: a geometrical-topological Total Variation Energy (*TVE*) index based on two-dimensional Minkowski functionals, designed to enhance sensitivity to localized anomalies. The proposed framework is validated experimentally using pulse-heated IRT data acquired from a carbon fiber-reinforced polymer (CFRP) plate containing six artificial defects at depths between 0.135 mm and 0.810 mm and is further supported by one-dimensional N-layer thermal model (NLM) simulations. The results demonstrate that the proposed metrics enable robust, unbiased ranking of image sequences and provide a reliable basis for automated defect-oriented image selection in IRT.

*Keywords:* composite materials, non-destructive testing (NDT), infrared thermography (IRT), representative elementary volume (REV), image analysis, homogeneity index (HI)

## 1. Introduction

The reliable and efficient non-destructive testing (NDT) of composite materials is of increasing importance, as material defects can critically affect the structural integrity of components. In this context, infrared thermography (IRT) provides a rapid, non-contact approach for detecting thermal anomalies potentially indicative of hidden defects. A range of post-processing methodologies has been developed to enhance the contrast of defect signals from thermograms, including, but not limited to, the following: Pulse Phase Thermography (PPT) [1], Thermographic Signal Reconstruction (TSR) [2], Principal Component Thermography (PCT)

[3], Differential Absolute Contrast (DAC) and its modifications (MDAC) [4-6], Thermal Shock Response Spectrum (TSRS) [7-9], and many more. The IRT experiment itself, like, generates large amounts of data. During an IRT experiment, usually a time series of thermograms is recorded for each sample. In PPT, the data is then transformed from the time- into the frequency domain using a discrete Fourier transform (DFT). This process yields amplitude and phase spectra, whereby a characteristic penetration depth can be interpreted based on its corresponding frequency. In TSR, the cooling curves of the thermograms – excluding the saturation phase [10] – are fitted pixel-wise with an $n$-th degree polynomial, and only the resulting coefficients are stored. Despite the above mentioned differences in the specific processing approaches, the examples illustrate a general principle: a significant number of IRT post-processing methods, whether explicitly or implicitly, generate structurally similar datasets in the form of images. They are typically represented as a three-dimensional image sequence, either with time, frequency, or a series of coefficients as the third dimension. These can then subsequently be analyzed to detect and characterize defects in the examined specimen. Figure 1 schematically illustrates this general situation. In practice, defect analysis leads to a key challenge: namely, determining which image within the sequence provides the most reliable representation of the defect, i.e., exhibits the highest defect contrast. The evaluation of the effectivity of different IRT data evaluation methods is typically carried out using specimens with integrated, well-defined artificial defects, for which parameters such as size, position, depth, and material are known. In this context, quantitative assessment and comparison are often based on metrics such as the signal-to-noise ratio (SNR) [11] and the Tanimoto criterion (TC) [12]. The Tanimoto criterion is a dimensionless parameter that enables a fast and objective comparison of images or other data. It involves information about correctly and incorrectly identified, as well as undetected, defect pixels. Owing to this property, it can be effectively used to assess the performance of different signal post-processing techniques and to compare them with one another regarding their capability to detect hidden defects in the depicted specimen. However, the accuracy of the Tanimoto criterion depends critically on the quality of the defect detection or contour search algorithm, since even minor inaccurately detected boundaries of defects can significantly affect the calculated result. The SNR is defined in various ways depending on the scientific discipline and the practical context of its application [13-15]. In infrared thermography, it provides a measure of the contrast between the defective area (region of interest) and a corresponding reference region. It is noteworthy that the measured SNR can be influenced by factors such as lateral heat flow, uneven heating, and also by the selection of the reference region. These factors may affect the absolute SNR values and the quantitative assessment of defect contrast. However, metrics that rely on the known position of a defect or on a carefully selected reference region are often of limited practical use, particularly in real-world scenarios and automated data processing, when the position of a potential defect is not known. In this paper, therefore, we propose a methodology that enables the identification of images within a dataset that are most likely to contain and represent a defect. This approach is of special relevance in scenarios where the existence of a defect is unknown, as it enables an unbiased, data-driven selection of images with the greatest likelihood of revealing anomalies. The approach provides a robust framework for automated processing and reliable decision-making in infrared thermography datasets and beyond by decoupling the analysis from prior knowledge of defect location or reference selection. Moreover, a comparison and evaluation of the empirical results with the theoretical results is provided, based on one-dimensional thermal N-layer model (NLM) simulations [16]. For our purposes, we use three different metrics: the first is the Homogeneity Index of Mixture (*HI*), which quantifies statistical heterogeneity by measuring deviations of local intensity distributions from the global reference distribution. Conceptually, this measure is related to histogram-based homogeneity measures and distributional criteria used in image analysis [17-19] as well as to classical mixing indices from mixture theory, where spatial uniformity is described through distributional metrics [20-22].The second metric used is derived from a widely used methodology called Representative Elementary Volume (*REV*) analysis [23, 24].

More specifically, it is based on a variant of this methodology that uses morphological measures on porous media - i.e., the Minkowski functional - to evaluate the *REV* [25]. In general, the *REV* analysis method identifies the smallest region of a heterogeneous material that sufficiently represents its overall geometrical and physical properties, enabling the estimation of macroscopic behavior from a limited sample. We adapt the methodology to describe two-dimensional images using a numerical metric called Representative Elementary Area (*REA*). Building upon these approaches, we introduce a third metric that is specifically designed to respond more sensitively to anomalies within image sequences, as will be demonstrated in the following sections. This metric is a consolidated index based on two-dimensional Minkowski functional, referred to as the geometrical-topological Total Variation Energy (*TVE*) index. For this purpose, a CFRP plate containing six artificial defects integrated at varying depths (from 0.135 mm to 0.810 mm) was employed as the test specimen.

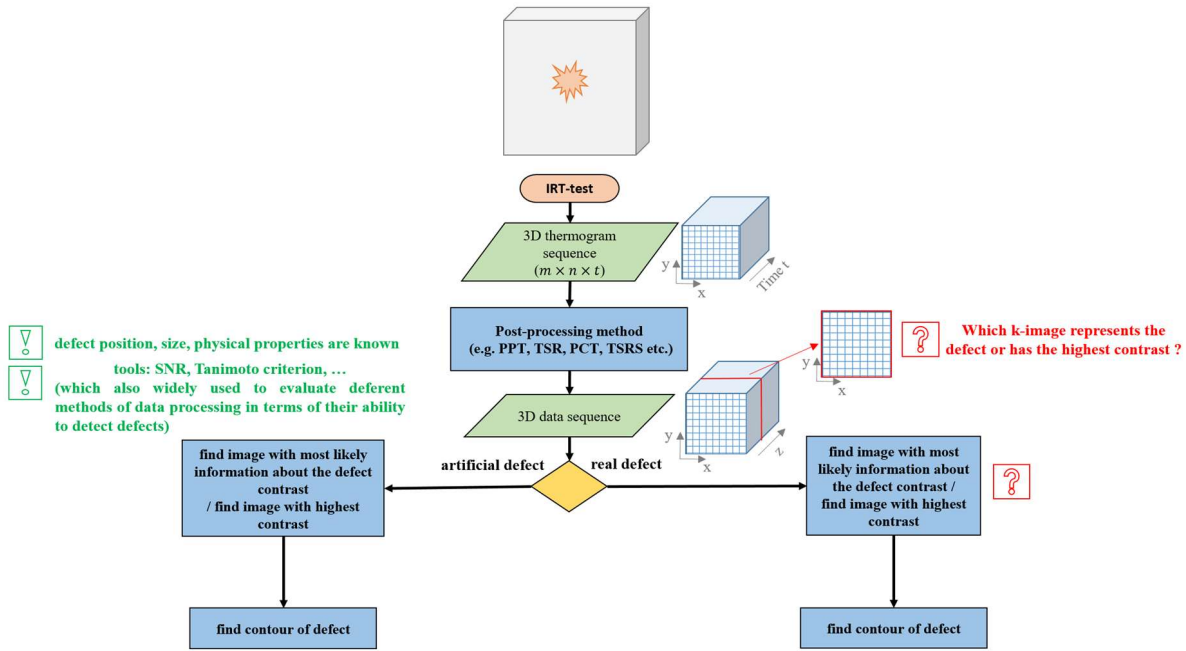

*Figure 1: The following schematic representation provides a visual flowchart of a standard thermographic testing, data acquisition, and processing workflow. he left tree of the figure illustrates the methodical evaluation process that employs known defects, whether artificial or integrated, with fully known parameters such as size, position, depth, and material, which are evaluated using suitable metrics (e.g., SNR, Tanimoto). The right branch represents the real-world application without prior knowledge of the existence or location of a defect or the most representative image within the sequence.*

## 2. Theory

### 2.1 Pulse Phase Thermography (PPT)

The development of Pulse Phase Thermography (PPT) was driven by the objective of integrating the strengths of both pulsed (IT) and modulated (LIT) infrared thermography. This integration involved isolating the various frequencies produced in pulsed infrared thermography to retain only the key phase components in the Fourier spectrum. This results in improved defect shape resolution, removes the requirement for pre-identifying defect-free areas within the field of view, and enables the inspection of high thermal conductivity materials [1]. In practice, the pixelwise discrete Fourier transform is applied to the discrete thermograms recorded by an infrared (IR) camera over a defined interval (from 0 to N) (1). Consequently, we obtain amplitude and phase sequences as a function of frequency $f$ instead of time-series thermograms. Such transformation leads to a transition to the frequency domain, resulting in an improvement in the contrast between defect and defect-free areas of a specimen. The amplitude and phase contrasts complement each other, and the phase contrast is independent of the reflectivity of the sample surface or of an uneven illumination during the impulsive heating process [26].

$$F(u) = \frac{1}{N} \sum_{n=0}^{N-1} f(n) \cdot e^{-\frac{j2\pi un}{N}} = Re(u) + j \cdot Im(u) \qquad (1)$$

where $Re(u)$ and $Im(u)$ are the real and imaginary parts of $F(u)$. The amplitude $A$ and phase $\Phi$ can be calculated as follows.

$$A(u) = \sqrt{(Re(u))^2 + (Im(u))^2} \qquad (2)$$

and

$$\Phi(u) = \tan^{-1}\left(\frac{Im(u)}{Re(u)}\right) \qquad (3)$$

## 2.2 Signal-to-noise-ratio (SNR):

The *SNR* is a reference-based measure that describes the contrast between the area of defect (signal) and the intact area (noise). It is described in many ways in literature [13-15]. We use it as in [27].

$$SNR = 20 \cdot log_{10}\left(\frac{|\bar{x}_{def} - \bar{x}_{ref}|}{\sigma_{ref}}\right) [dB] \qquad (4)$$

where $\bar{x}_{def}$ is the average signal in a defective area and $\bar{x}_{ref}$ in reference respectively. For normalization the standard deviation $\sigma_{ref}$ of the reference area is used.

## 2.3 Tanimoto Criterion (TC)

The Tanimoto criterion $T_c$ is a dimensionless measure used for rapid comparison of image data. It is defined as

$$T_c = \frac{N_{r,d} - N_{m,d}}{N_{r,d} + N_{f,d}} \qquad (5)$$

where $N_{r,d}$, $N_{f,d}$, and $N_{m,d}$ denote correctly detected, falsely detected, and missed defect pixels, respectively. In IRT, $T_c$ provides a normalized metric for evaluating and comparing defect detection methods [9,12].

## 2.4 Minkowski functional

[28] In two dimensions, the Minkowski functional $M_0$, $M_1$ and $M_3$ are quantitative measures that describe the morphological properties of a structure in terms of its area, perimeter, and Euler characteristic, which represents the topological connectivity of a structure, respectively.

They are defined as follows:

$$M_0(X) = \int_X ds \tag{6}$$

$$M_1(X) = \frac{1}{2\pi} \int_{\delta X} dc \tag{7}$$

$$M_2(X) = \frac{1}{2\pi^2} \int_{\delta X} \left[\frac{1}{R}\right] dc \tag{8}$$

where 2D-body $X$, with a smooth boundary $\delta X$, $ds$ is the surface element, $dc$ is a circumference element, and $R$ is the radius of local curvature.

## 3. Material, experimental setup and methodology

### 3.1  Specimen and experimental setup

A single CFRP panel was manufactured for testing using a unidirectional prepreg from Solvay Inc. (Belgium), composed of Teijin Tenax HTS40 carbon fibers embedded in CYCOM 977-2 epoxy matrix. The panel was cured in an autoclave under standard supplier conditions (7 bar pressure, 180 °C temperature for 180 min), resulting in a specimen of $300 \times 300\ mm^2$ with a nominal thickness of 1.7 mm and a quasi-isotropic layup of 13 layers $(45/-45/45/-45/90/0/\overline{90})_s$, representative of typical aerospace components [29]. Artificial defects were inserted as thin Tetrafluerenhylen-Hexafluorpropylen Copolymer (FEP) foils ($50\ \mu m$) placed at six different depths between prepreg layers (Figure 2). The defect size ($15 \times 15\ mm^2$) and positions were chosen to reduce edge effects and interactions, simulating realistic delamination scenarios, particularly in aerospace applications. FEP was selected to simulate regions where heat accumulation occurs due to interlayer separation, providing a representative case for thermal analysis. The thermophysical properties of the materials were determined using standardized measurement methods. The density of the CFRP panel was measured according to DIN EN ISO 1183, specific heat capacity according to DIN EN ISO 11357-4, and out-of-plane thermal conductivity according to DIN EN ISO 11357-8. Properties of the FEP foils were obtained from the manufacturer (Holscot Europe). For the surrounding ambient air, standard thermophysical properties were used. Values for density, specific heat capacity, and thermal conductivity were taken from the literature [30] and are summarized in Table 1.

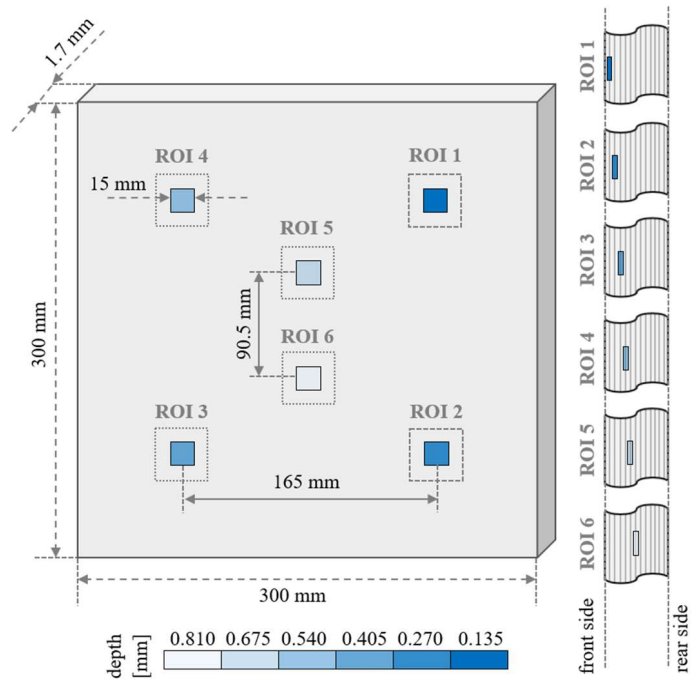

*Figure 2: Schematically representation of CFRP plate.*

Table 1: Physical properties of used materials and ambient air for simulation. For CFRP, the thermal conductivity is out-of-plane.

|  | CFRP | FEP | air |
|---|---|---|---|
| Density: $\rho$ [$kg/m^3$] | 1602 | 2200 | 1.204 |
| Specific heat capacity: $c_p$ [$J/(kg \cdot K)$] | 930 | 1145 | 1006 |
| Thermal conductivity: $\lambda$ [$W/(m \cdot K)$] | 0.35 | 0.23 | 0.026 |
| Thermal effusivity: $e$ [$J/(m^2 \cdot K \cdot \sqrt{s})$] | 722.11 | 761.16 | 5.61 |

In the IRT experiment, the external heat source was provided by two xenon flash lamps VH3-6000 with Tria 6000 S (Hensel GmbH, Germany) generators, each capable of delivering pulses ranging from 187 to 6000 J of energy. The duration of the impulse was set to 2 ms. The positioning of the lamps at an angle less than 30° relative to the specimen surface was implemented to prevent any potential influence on the blind frequency [31]. Additionally, a 9.27 mm Polymethylmethacrylate (PMMA) plate was positioned in front of each lamp, serving to suppress near-infrared radiation within the operating range of the IR camera [32, 33] and thereby ensure the accuracy ad quality of the thermographic measurements [32]. Surface temperature measurements were obtained using an IR camera (IR9410BI with max. 1280 x 1024 pixels resolution and thermal sensitivity (NETD) of 20 mK at 30 °C), with a pixel resolution of 0.3 mm and frame rates 180 frames per second. An active photodiode (PDA36A2, Thorlabs, USA), equipped with suitable neutral density filters (ND) from Thorlabs, was implemented to monitor the flash pulse. This approach was adopted to account for the occurrence of cutting afterglow effects from Thermograms in the pre-treatment stage [34]. Figure 3 shows the experimental setup and its schematic representation.

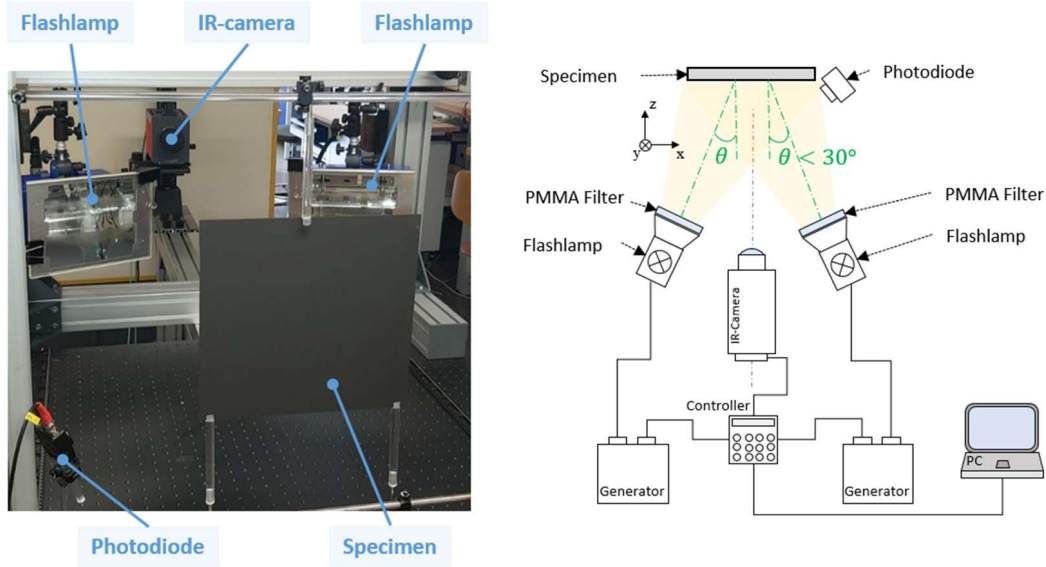

Figure 3: The experimental setup for the active IRT experiment is shown on the left, using two xenon flash lamps. The corresponding schematic representation of the experimental setup can be seen on the right.

## 3.2 Methodology

### 3.2.1 Evaluation methodology

The proposed metrics - formulated in detail in the preceding sections - are evaluated for their suitability using the following procedure. Each image sequence is treated as an index-ordered set of frames, and the metric value is computed for every image index. This results a one-dimensional metric curve that represents the evolution of the structural characteristics captured by the metric. In this context, the notion of suitability refers to how well the resulting metric curve reflects the presence, persistence, and evolution of defect-related structural patterns within the sequence. For comparative assessment, the signal-to-noise ratio (SNR) and the Tanimoto criterion are employed as reference metrics, allowing the proposed measures to be evaluated against established indicators of image quality and structural relevance. The calculation of the SNR, as per Equation (4), for each integrated defect in the plate illustrated in Figure 2, the reference region was selected as follows. A region of interest (ROI) containing the defect was initially specified from the entire dataset. Within this ROI, the area surrounding the defect was designated as the reference region (Figure 4), thereby providing a basis for calculating the SNR relative to the defect. The determination of the defect region and the subsequent calculation of the Tanimoto criterion are achieved through the implementation of the contour-search algorithm, as outlined in [9]. This method has demonstrated an enhanced performance in comparison to conventional approaches, particularly in the context of automated processing of IRT datasets [9]. In response to the potential impact of factors such as non-uniform heating during the experiment on both the ROI and the computed metrics - particularly the two reference metrics, SNR and TC - a spatial filter is applied to each image during the preprocessing phase. A detailed description of the filter, including its advantages and limitations, is provided and discussed in [8]. The filter's operation involves the subtraction of an empirically interpolated image, which is generated according to the formula (9), from the original image. This process reduces gradual intensity variations and enhances the contrast of local defect-related features.

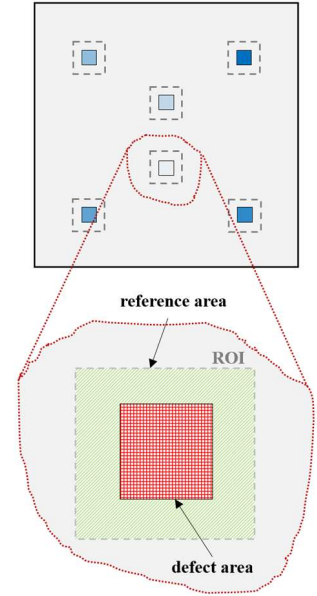

*Figure 4: The schematic representation of ROI, reference (marked in green), and defect (marked in red) area.*

$$z = f(x, y) = c_1 x^2 + c_2 y^2 + c_3 \tag{9}$$

where $c_1$, $c_2$, and $c_3$ are interpolation parameters.

### 3.2.2 Homogeneity Index of Mixture (*HI*)

A variety of established methods exist for evaluating the spatial homogeneity of image data. These approaches typically rely on subdividing an image into macroscopic subregions (macropixels), from which local statistical descriptors are derived. Several homogeneity criteria have been proposed in the literature, including macropixel-based indices originally developed for assessing blending processes in solid particulate systems [17], and distribution-based methods such as the Distributional Homogeneity Index (DHI), which compares local macropixel statistics with the global intensity distribution [19]. Further studies have extended these concepts by adjusting subregions sizes or introducing alternative measures to quantify distribution similarity [35, 36]. In addition to hyperspectral imaging approaches, the theoretical foundations of quantitative homogeneity assessment are also grounded in classical mixing theory. Mixing indices developed for solid particles and pharmaceutical blends [18, 20-22] describe the transition from ordered to statistically mixed states based on spatial distributions, variance reduction, and macroscopic uniformity. While these approaches originate from different domains (particulate systems vs. imaging), they share the same conceptual principles of variance normalization, stochastic structure characterization, and spatial uniformity assessment. These classical mixing models therefore provide a complementary theoretical framework supporting the conceptual development of the Homogeneity Index of Mixture (*HI*) defined in this paper, highlighting its basis in well-established statistical and mixing-theory concepts.

The *HI* proposed herein is a metric quantifying the homogeneity of pixel distributions within an image. At first the image will be normalized to the interval [0,1]. Subsequently, the image is divided into $n$ cells of the size $MxM$. Each cell $j$ must contain at least as many pixels as there are bins $k$ in the distribution for the data to be considered valid. For the whole image the reference probability $p_i$ is calculated using:

$$p_i = \frac{c_i}{c}, i \in [1; k] \tag{10}$$

To determine the HI, the variance $\sigma_i^2$ of the bin-wise probability values across all $n$ cells is calculated:

$$\sigma_i^2 = \frac{1}{n}\sum_{j=1}^{n}(p_{ij} - p_i)^2 \tag{11}$$

The *HI* is defined as the sum of the square root of the variance for every bin:

$$HI = \sum_{i=1}^{k}\sigma_i = \sum_{i=1}^{k}\sqrt{\frac{1}{n}\sum_{j=1}^{n}(p_{ij} - p_i)^2} \tag{12}$$

Calculation process can be executed in either a static or dynamic manner. In static mode, the image is divided into a fixed grid. In contrast, in dynamic mode, the cells are selected randomly with each iteration, and the *HI* is calculated. The calculation is completed when the *HI* converges. The optimal homogeneous image is characterized by a *HI* of 0.

### 3.2.3 Representative Elementary Area (*REA*) & Total Variation Energy (*TVE*)

The following section describes the procedure used to calculate the two metrics introduced, *REA* and *TVE*. As previously stated, the conceptual foundation of our approach is based on the Representative Elementary Volume (*REV*) employing Minkowski functionals.

In the first stage (Figure 5), each image $I$ from the 3D data sequence is empirically segmented into n phases/clusters ($\Phi = 1, 2, \cdots, \phi$). To characterize local morphological features, square windows of size $nxn$ are evaluated over the segmented image. The window size $n$ is systematically varied within a predefined range $n \in \{n_{min}, \ldots, n_{max}\}$. Typically starting from $n_{min} = 2$ to $n_{max}$, which is constrained by the image dimensions. It is important to note that it is advisable to select the minimum size so that the number of pixels in the selected minimal window is equal to or greater than the predefined $\phi$-phases in the hole image $I$. For each window, the three standard Minkowski functionals are computed. Two conceptually different sampling strategies can be used to place the windows: deterministic and statistical (Figure 6). Although deterministic, center-based window expansion or estimation of REV [25] has been reported in the relevant literature, its applicability here is limited or unfavorable.

Therefore, only statistical sampling strategies are considered. Two variants of statistical sampling are employed and considered to generate representative local morphological measurements. In the static variant, the image is partitioned into a regular, non- overlapping grid of windows with size $nxn$. For each grid cell, the Minkowski functionals $M_0$, $M_1$, and $M_2$, are computed. This procedure ensures full spatial coverage but remains constrained by the imposed grid structure. In the random variant, window positions are drawn uniformly at random from the image domain. A predefined number of samples (NOS) $k$ of non-identical windows that may partially overlap is selected for each window size $nxn$. For both static and random sampling, a representative generalized measure of the Minkowski function-based metrics at each window size is obtained using the normalized coefficient of variation ($CV_{norm}$) [37, 38].

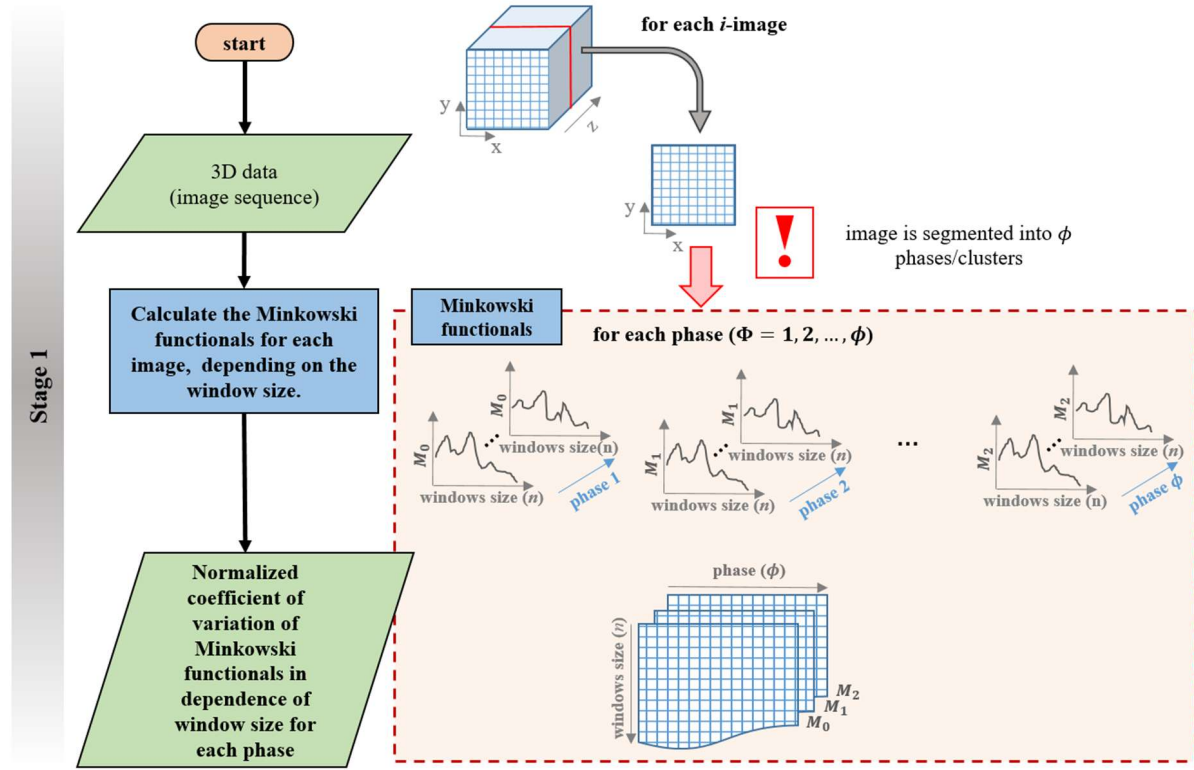

*Figure 5: The following flowchart illustrates Stage 1 of the proposed calculation methodology. Initially, each image I from the 3D data sequence is empirically segmented into n phases/clusters (Φ=1, 2, ⋯, ϕ). For each phase and window size n, the Minkowski functionals $M_0$, $M_1$, and $M_2$ are computed. Subsequently, the normalized coefficient of variation of the Minkowski functionals is determined as a function of the window size for each phase.*

$$CV_{norm} = \frac{\sigma/\mu}{\sqrt{k}} \qquad (13)$$

where $\sigma$ and $\mu$ denote the mean and standard deviation, respectively, of the Minkowski functionals computed across all windows of size $nxn$. The employment of the normalized coefficient of variation ($CV_{norm}$) is particularly advantageous, as the amount of data contributing to the statistical evaluation varies with several factors, such as the overall image size, the selected sampling strategy, and the number of available windows for each size n. Normalization effectively compensates for these variations and ensures that the heterogeneity measure remains comparable across different window sizes and sampling configurations

In the second stage (Figure 7), a Topological-Geometrical Index (TGI) is defined based on Minkowski functionals by summing all three calculated Minkowski functionals. In addition, the TGI-derivation's dependence on the window size is determined for each phase. This provides the foundation for further processing.

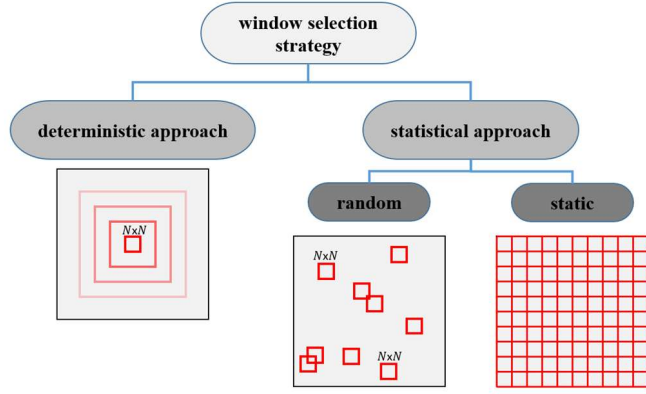

Figure 6: Schematic overview of the window selection strategy. A deterministic approach is contrasted with statistical approaches, which include random and static window selection.

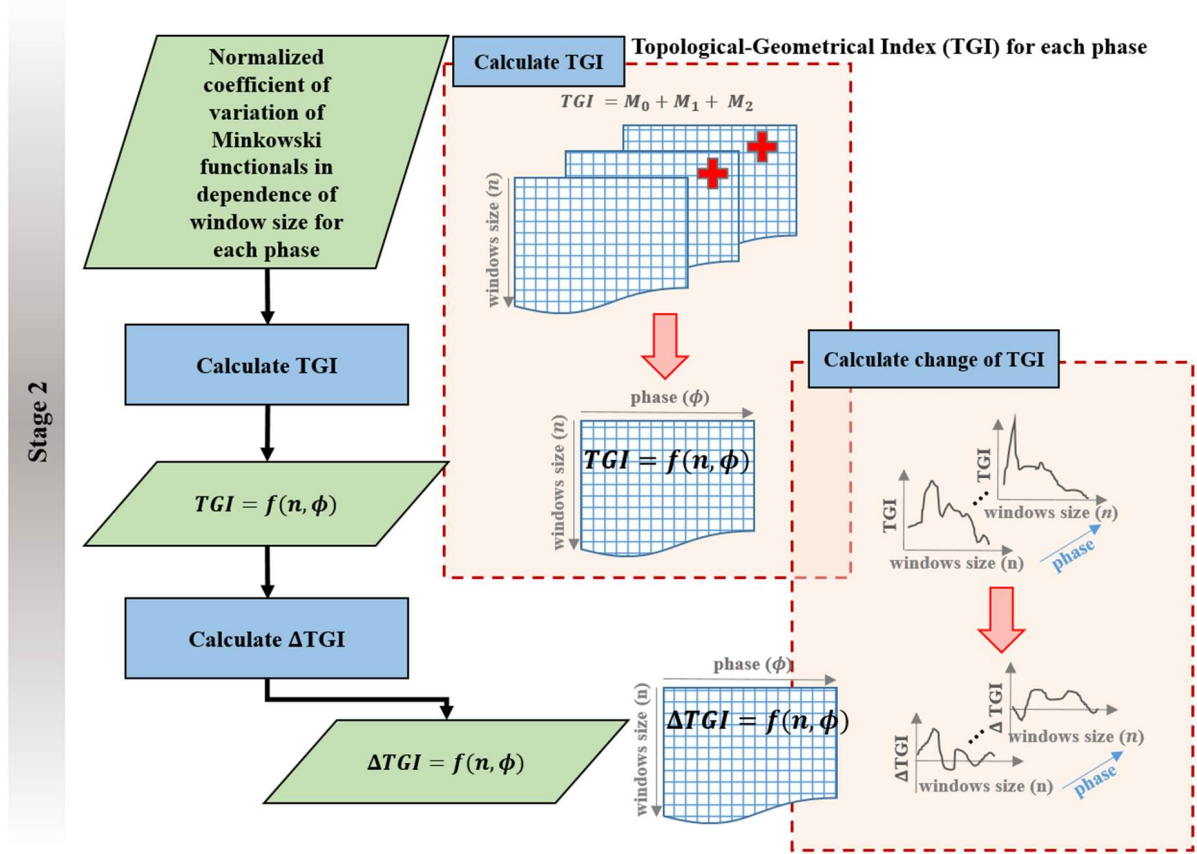

Figure 7: The following flowchart illustrates Stage 2 of the proposed calculation methodology. The calculation of the Topological Geometrical Index (TGI) for each phase is determined by summing all three calculated Minkowski functionals. The TGI-derivation's dependence on the window size is additionally determined for each phase.

In the third stage (Figure 8), we define two different metrics based on the $\Delta TGI = f(n, \phi)$ curves. The first metric, Total Variation Energy ($TVE$), is defined as follows:

$$TVE = \sum_{\phi} \sum_{n} (\Delta TGI^2) \qquad (14)$$

To determine the representative elementary area ($REA$) of Image $I$, the $\Delta TGI$ curves are evaluated separately for each phase. The window size at which each curve reaches convergence is identified for each phase and interpreted as the phase-specific $REA$. If a $\Delta TGI$ curve does not converge within the predefined range of window sizes, the largest possible window size $n_{max}$ is assigned as the $REA$ for that phase. To obtain a generalized $REA$ for the entire image, the maximum of all the phase-specific $REA$ values is selected to ensure that the resulting metric reliably captures the structural characteristics across all phases. In both cases, each image provides a single scalar value, enabling the construction and analysis of a corresponding curve over the entire sequence. It is expected that images containing a defect or exhibiting a local inhomogeneity will produce higher metric values, typically manifested as pronounced local maxima within the sequence.

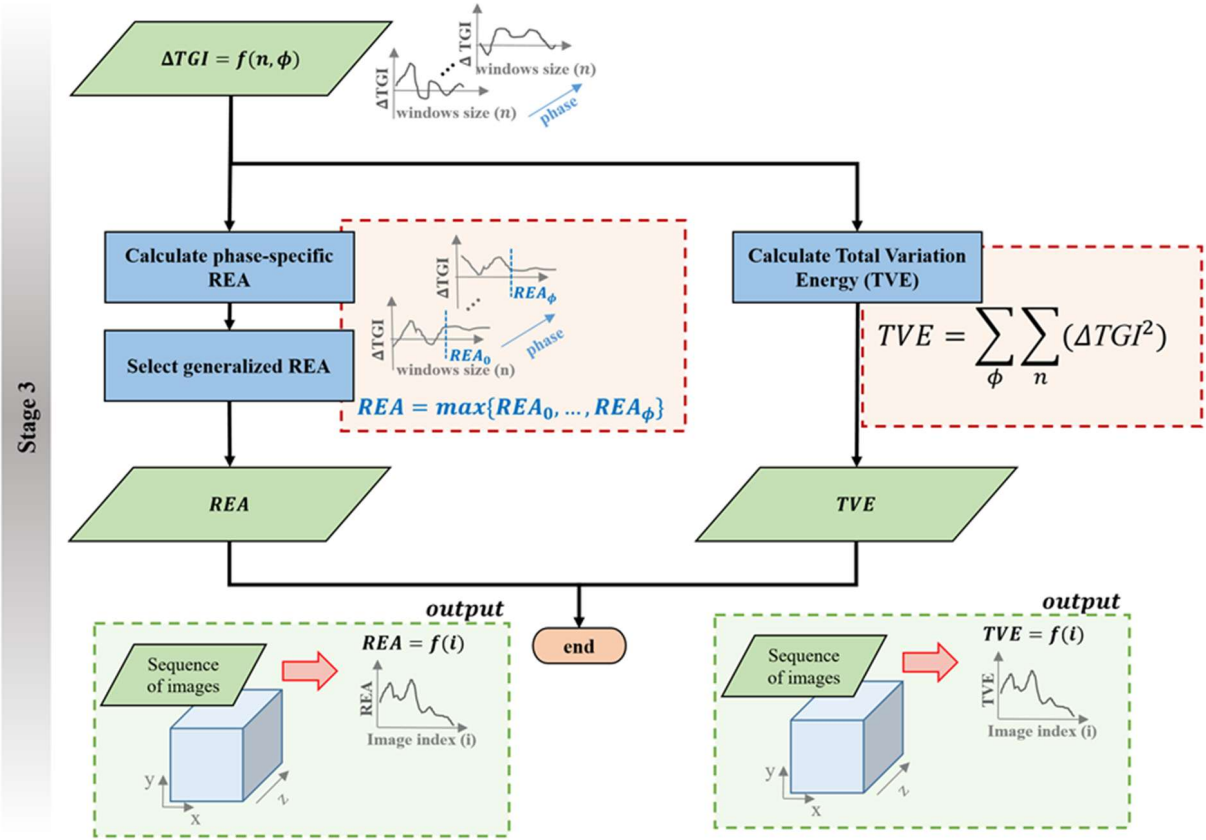

*Figure 8: The following flowchart illustrates the last Stage 3 of the proposed calculation methodology. The final results are to be calculated in terms of index-dependent REA and TVE curves.*

## 4. Results and Discussion

All methods and metrics described in Chapter 3, as well as the conducted investigations, were fully implemented in the Python environment.

A salient aspect in the implementation of the $HI$ metric is the selection of the number of bins used for probability estimation. Given that all images in the sequence possess identical dimensions and, consequently, an equivalent number of observations per image, the number of bins is determined based on this number. According to [37, Chapter 3.5.1], the number of bins $k$ can be roughly selected approximately to the square root of the number of observations $n$ in an image ($k \sim \sqrt{n}$). For images with a very large number of observations ($n \geq 1000$), a logarithmic scaling is recommended ($k \sim 10 \cdot \log_{10}(n)$). This approach ensures an adequate representation of the data distribution while enabling a fully automated calculation of the $HI$ metric.

Another part in of the workflow benefits significantly from automated calculation, which makes the entire process more efficient and structured. This is particularly relevant in determining the onset of the convergence range for phase-specific REA curves, as discussed in section 3.2.3. To determine the onset, a method widely used in signal processing, namely the

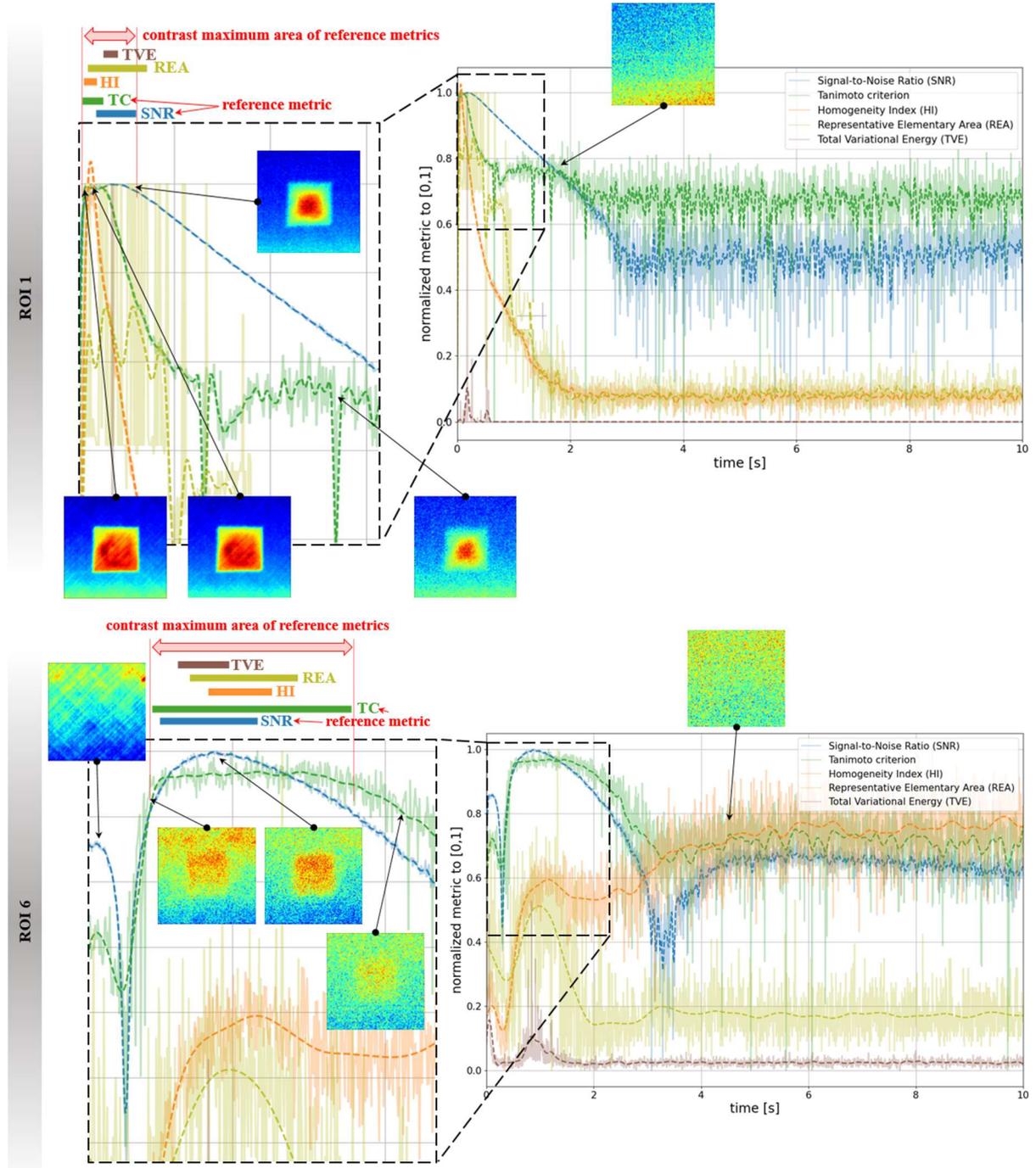

*Figure 9: The reference and proposed metric curves are shown as functions of the sequence index (for thermogram sequences, the index corresponds to time). The phase numbers were set at 4 and NOS at 439 for the calculation of REA and TVE. Given the marked variation in numerical ranges across the metrics, a normalization of all curves to the interval [0,1] was employed. The results for ROI 1 are displayed in the upper diagrams, while those for ROI 6 are shown in the lower diagrams. Additionally, regions containing local and global maxima are magnified. Due to the presence of significant noise in the signals, the curves were filtered with low-pass Butterworth filter, and they are displayed as dotted lines to enhance visual clarity. Furthermore, bars are employed to denote index ranges corresponding to locally/globally maximum metric value ranges, where the bar colors correspond to the colors of the respective metric curves. The corresponding images were displayed at selected locations or indices (the images without filtering)*

Akaike information criterion (AIC) [39], was employed. The application of AIC has proven to be effective not only in the domain of acoustic emissions [40], but also in other data analysis applications [41].

During the experiments, a xenon lamp was used to thermally excite the entire CFRP plate, enabling full-field thermographic acquisition of the specimen surface. Regions of interest (ROIs) were then extracted from the recorded thermograms and processed as image sequences. During pre-processing, saturated regions [10] as well as frames affected by the afterglow effect [34] were excluded from further analysis. The thermographic images had a pixel size of approximately 0.3 mm, resulting in ROI dimensions of 118 × 118 pixels. Initially, the proposed evaluation methodology described in Section 3.2.1 was applied to the cropped ROI thermogram sequences. To reduce computational time, the analysis was limited to the first 10 seconds of each thermogram sequence. Two representative results corresponding to the near-surface and deepest defect regions, denoted as ROI 1 and ROI 6, respectively, are presented in Figure 9. The analysis of the remaining ROIs is provided in Appendix A.

A comparative analysis between the three proposed metrics (*HI / REA / TVE*) and the reference metrics (SNR / TC) reveals that they generally exhibit qualitatively similar behaviors. In regions of high defect detectability - corresponding to indices with pronounced contrast - they display well-defined global or local maxima, indicating clear sensitivity to defect areas. Conversely, in regions of low defect detectability at higher indices, all proposed metrics converge towards a nearly constant value, reflecting stable baseline behavior. Across all metrics, with the exception of *TVE*, noise levels are relatively high and largely independent of the ROI. In order to enhance visual clarity and facilitate comparison, high-frequency components were suppressed using a band-pass filter, and filtered signals are indicated with dashed lines in the figures. The behavior of *TVE* differs significantly in regions with low error detectability. In such regions, *TVE* is very close to zero and exhibits nearly linear convergence. This underscores the instrument's inherent robustness and minimal sensitivity to background noise in these regions, thereby establishing a sharp and interpretable quantitative baseline.

As previously mentioned in section 3.2.3, only statistical selected windows strategies are taken into account for the calculation of morphological measurements. The results of the investigations indicate that, despite the computational efficiency of the static window strategy, its inherent methodological constraint - namely the use of a rigid grid - limits its ability to consistently achieve the objectives defined in this paper. Consequently, a methodology referred to as random (randomized) was employed to select the windows in this study. The selection of NOS is a crucial factor in this approach, as the consideration of all possible windows results in a significant increase in computing effort, thereby significantly affecting the computing power and practical applicability of the method. The CV of the Minkowski functionals has been computed and utilized as the basis for the evaluation of the TVE and REA metrics. Therefore, the preliminary determination of NOS was made through the implementation of a sample size planning for the CV approach, with the methodology proposed in [42]. In accordance with the parameters specified in Table 3 [42], the number of NOS was set to 439, with a coefficient of variation of 0.5, a target confidence interval width of $w = 0.125$, and a degree of assurance $\gamma = 0.99$ of achieving the desired confidence interval. The assumption that CV = 0.5 is a conservative one for the variability expected in thermographic measurements. This assumption ensures a sufficiently robust sampling. It is important to note that when a constant number of NOS is selected ($NOS_{set} = const$), the effective selection of window samples is influenced by two factors: the overall image size and the dimensions $n \times n$ of the selected window. For certain window sizes $n \times n$, the total number of possible window samples may be less than the predefined number of $NOS_{set}$. In such cases, all possible windows are evaluated completely. However, when the total number of possible windows exceeds $NOS_{set}$, the corresponding number of windows is selected randomly. Figure 10 schematically illustrates this behavior and contrasts the random and static NOS selection strategies as a function of window size. In addition to the NOS determination strategy described above, a further variability analysis was performed in which both the number of NOS values and the number of calculation iterations (NOI) were systematically varied in order to evaluate their effects on the final results. The investigations show a clear correlation with the NOS setting strategy and demonstrate convergence in many cases even at NOS values below 439, especially for the indices of local and global maxima proposed metrics, as well as the size of the corresponding confidence intervals. This finding indicates that a reduced number of NOS values may be sufficient to obtain stable and representative results, further reducing computational costs. This behavior of convergence is demonstrated using a representative example of the *TVE* metric for ROI 1 in Figure 11.

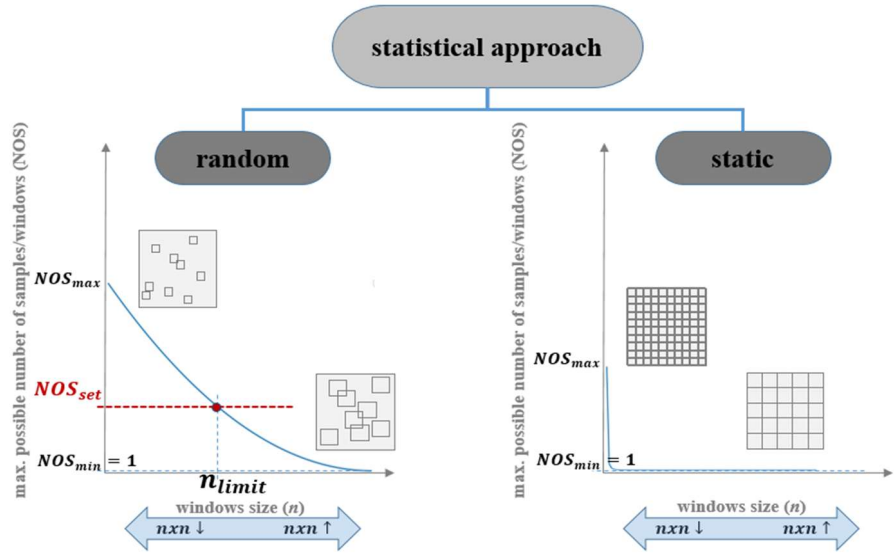

*Figure 10: Schematic comparison of two statistical window selection strategies: the static approach on the left and the random approach on the right, including an additional red marker indicating the case of constant NOS.*

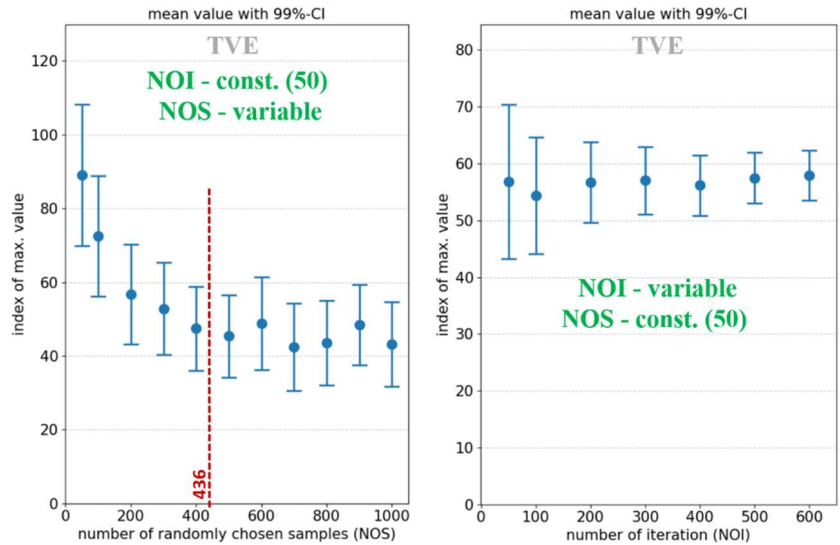

*Figure 11: the figure displays representative examples by ROI 1, which demonstrate the effect of the number of randomly selected window samples (NOS, left) and the number of iterations (NOI, right) on the results, particularly on the mean value of the maximum index of the TVE metric. The mean values of the maximum index are plotted with 99% confidence intervals, calculated according to the methodology described in Chapter 6.8.2 [37]. In the left-hand figure, the number of $NOS_{set} = const.$ is indicated by a red dotted line and was set to 439, based on the parameters specified in Table 3 [42].*

The proposed methodology, when applied to both amplitude and phase sequences derived from PPT data, yields highly comparable results. A representative example is shown in Fig. 12. Nevertheless, two phenomena can be identified with greater clarity. First, as previously discussed, minor discrepancies between the results obtained from the reference metrics may arise from the different computational principles underlying the SNR-based and Tanimoto-based approaches. Consequently, the two reference metrics may, on occasion, yield slightly different responses, with local and global maxima occurring at marginally shifted index ranges. Second, for deeper defects, two pronounced local maxima regions can be observed, in some cases exhibiting very similar magnitude values. This behavior is most likely attributable to the superposition of reflected signals.

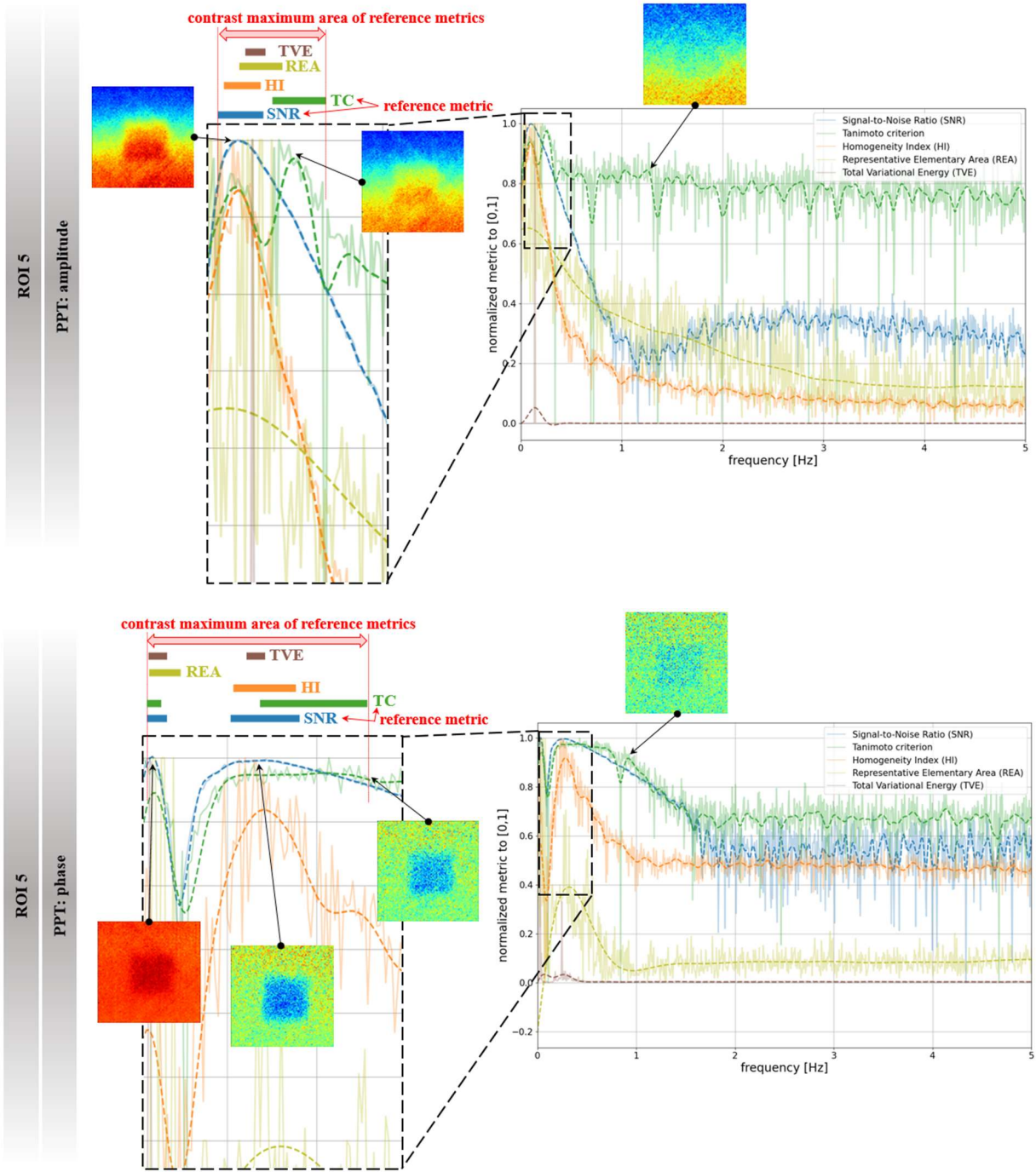

*Figure 12: The reference and proposed metric curves are shown as functions of the sequence index (frequency). The cluster numbers were set at 5 for amplitude and 4 for phase sequences and NOS at 439 for the calculation of REA and TVE. Given the marked variation in numerical ranges across the metrics, a normalization of all curves to the interval [0,1] was employed. The results for ROI 5 amplitude are displayed in the upper diagrams, while those for phase sequence are shown in the lower diagrams. Additionally, regions containing local and global maxima are magnified. Due to the presence of significant noise in the signals, the curves were filtered with low-pass Butterworth filter, and they are displayed as dotted lines to enhance visual clarity. Furthermore, bars are employed to denote index ranges corresponding to locally/globally maximum metric value ranges, where the bar colors correspond to the colors of the respective metric curves. The corresponding images were displayed at selected locations or indices (the images without filtering). Further results are provided in Appendix B.*

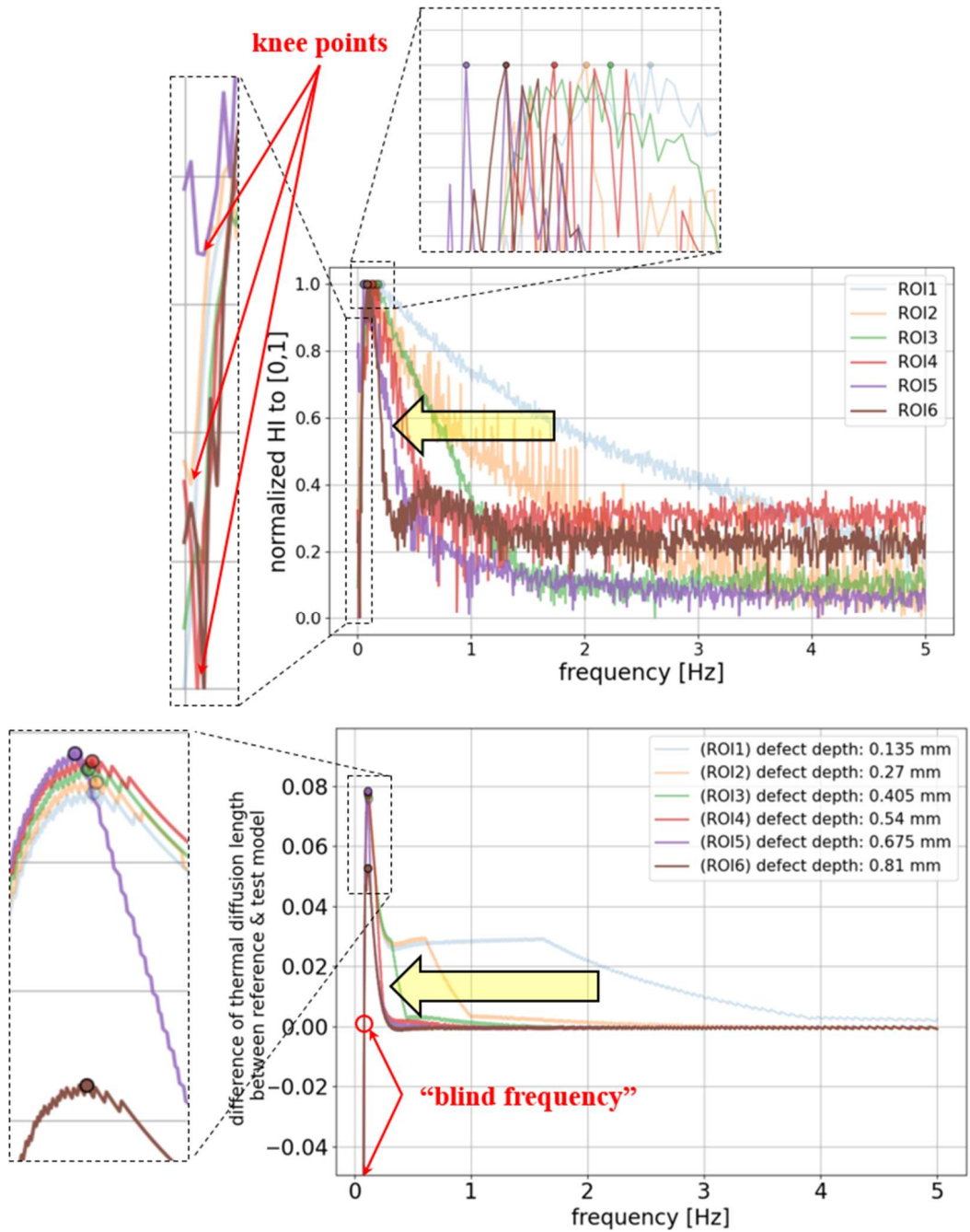

*Figure 13: Frequency-dependent HI curves (upper) and the corresponding difference in thermal diffusion length between reference and defect regions (bellow) for six ROIs with varying defect depths. The maximum points of each region are represented by corresponding colored dots. The trend of the curves is indicated by light yellow arrows from ROI 1 to ROI 6. A comprehensive graphical representation of the methodology employed for the generation of simulated curves is provided in Appendix C.*

A further method of evaluating the validity of the proposed methodology and the resulting findings is to compare these with theoretical predictions. To this end, a one-dimensional analytical solution of an N-layer theoretical model [16] was employed as a reference framework. Previous studies have shown good agreement between this analytical model and experimental infrared thermography (IRT) measurements, supporting its suitability for comparison with experimental results [8, 9, 43]. Furthermore, numerical simulations are used to estimate the thermal diffusion length [44], with particular focus on the difference between reference (1-layer model) and defect regions (3-layer model). The resulting contrast curves can be interpreted as indicators of defect detectability as a function of frequency. For instance, when considering the frequency-dependent *HI* curves obtained from amplitude images of PPT experiments for six different defect regions and comparing them with the corresponding simulations, namely the contrast curves based on the thermal diffusion length, several qualitative correlations become apparent (Figure 13). First, it can be observed that the contrast peaks for all ROIs are located very close to each other in both diagrams and occur in the vicinity of 0.1 Hz. Furthermore, the width of the peak region becomes progressively narrower, and the slope of the curves becomes steeper from ROI 1 to ROI 6, as illustrated with a slightly yellow arrow in Figure 13. Another notable observation concerns the frequency at which the difference in thermal diffusion length becomes zero, which is found to be very similar for all ROIs. The corresponding frequency, or frequency range, may be interpreted as a "blind frequency" region, in which defect detectability is not expected. A similar behavior is observed in the experimental data, where the *HI* curves show a comparable transition at lower frequencies, marked as the "knee point" in Figure 13. This phenomenon can be explained by the fact that the *HI* is incapable of assuming negative values, resulting in a significant change in slope at this particular frequency. Furthermore, an increase in contrast is observed beyond this point, which qualitatively correlates with the results of the simulations.

## 5. Conclusion

All methods and metrics proposed in this paper and introduced in Chapter 3 were implemented in full and systematically evaluated within a Python-based processing framework. Particular attention was given to the practical aspects of automation, robustness, and computational efficiency, which are essential for large-scale thermographic data analysis. The results demonstrate that the proposed metrics (*HI*, *REA*, and *TVE*) and evaluation methodology closely reproduce the qualitative behavior of the reference metrics (SNR and TC) as a function of the sequence index. This consistency is observed independently of whether thermogram sequences or phase- and amplitude-sequences are considered. Across all investigated cases, the proposed approaches reliably capture the characteristic trends, extrema, and convergence behavior exhibited by the established reference metrics. All metrics consistently reveal pronounced local or global extrema in regions of high defect detectability, while converging toward stable baseline values in regions of low detectability. Among the investigated approaches, the *TVE* metric demonstrates a distinct behavior characterized by near-zero values and nearly linear convergence in low-contrast regions, thereby highlighting its robustness and reduced sensitivity to noise. The noise levels in both the *HI* and *REA* curves are also high, similar to those observed in the reference metrics. In the case of the *REA* metric, this high noise can, in certain instances, not only reduce the interpretability of the curves but also render them effectively uninterpretable. The study further confirms that window selection strategies (Figures 6 and 10) play a crucial role in the stability and reliability of morphological measurements necessary for the calculation of *REA* and *TVE* metrics. While static window strategies offer computational efficiency, their inherent rigidity limits their general applicability. Conversely, the randomized window selection approach, in combination with statistically justified number of samples planning, offers a flexible and robust alternative. The results demonstrate that stable and representative outcomes can often be achieved with a reduced number of window samples, indicating significant potential for further reductions in computational cost without compromising result quality. A comparison of computation times among the proposed metrics reveals that HI generally requires significantly less processing time compared to *REA* and *TVE*. To illustrate, the processing of a single ROI sequence of dimensions 118 × 118 × 1781 currently requires approximately 47 seconds for the *HI* metric curve, depending on the available computing resources (see Appendix D). Nevertheless, there is considerable potential for optimization in future implementations, which could reduce the computation time of *REA* and *TVE*. Furthermore, the validity of the proposed methodology was further supported through comparison with theoretical predictions derived from a one-dimensional N-layer analytical thermal model, as well as with corresponding simulations and calculation of the thermal diffusion length. Especially, qualitative correlation was observed between the frequency-dependent *HI* curves obtained from the PPT amplitude image sequences and the corresponding simulated contrast curves based on thermal diffusion length. The findings of both approaches indicated that there was close alignment of peak frequencies of approximately 0.1 Hz for all regions of interest (ROIs), in addition to consistent trends in peak width and slope for defects ranging from the surface to the depths. In addition, a comparable transition region was identified in both the experimental and simulated results. This region is interpreted as a blind-frequency or knee-point region with reduced defect detectability. These findings confirm the ability of the proposed metrics to identify defect-related features and to correlate with simulated physical contrast predictions. Overall, the results confirm that the proposed metrics and framework provide a robust, automated, and computationally efficient system for detecting and further quantifying anomalies and defects in various thermographic data sequences, with great potential for extension to other materials, excitation schemes, and imaging methods.

# Appendix A

The subsequent appendix presents the corresponding metric curves for the remaining regions of interest (ROIs), shown analogously to Figure 9. It is to be noted that all curves are normalized to the interval [0,1], and the same processing parameters are applied (phase number = 4, NOS = 439). The filtered metric curves are displayed as dotted lines, while bars indicate index ranges associated with locally and globally maximum metric values.

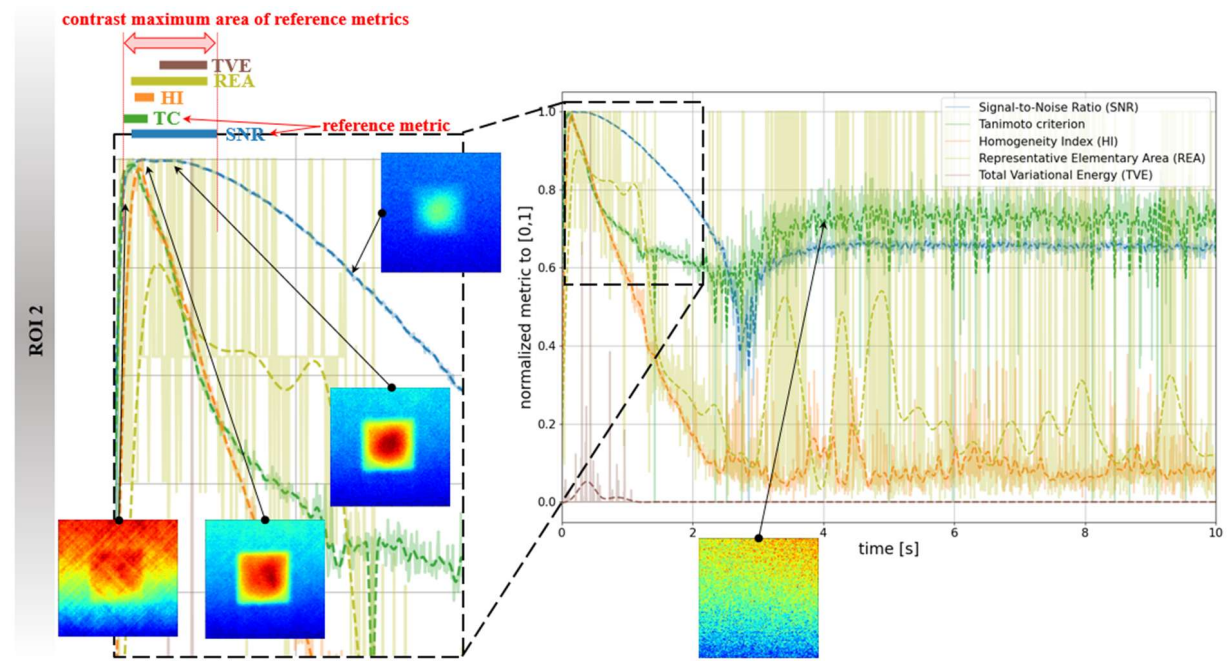

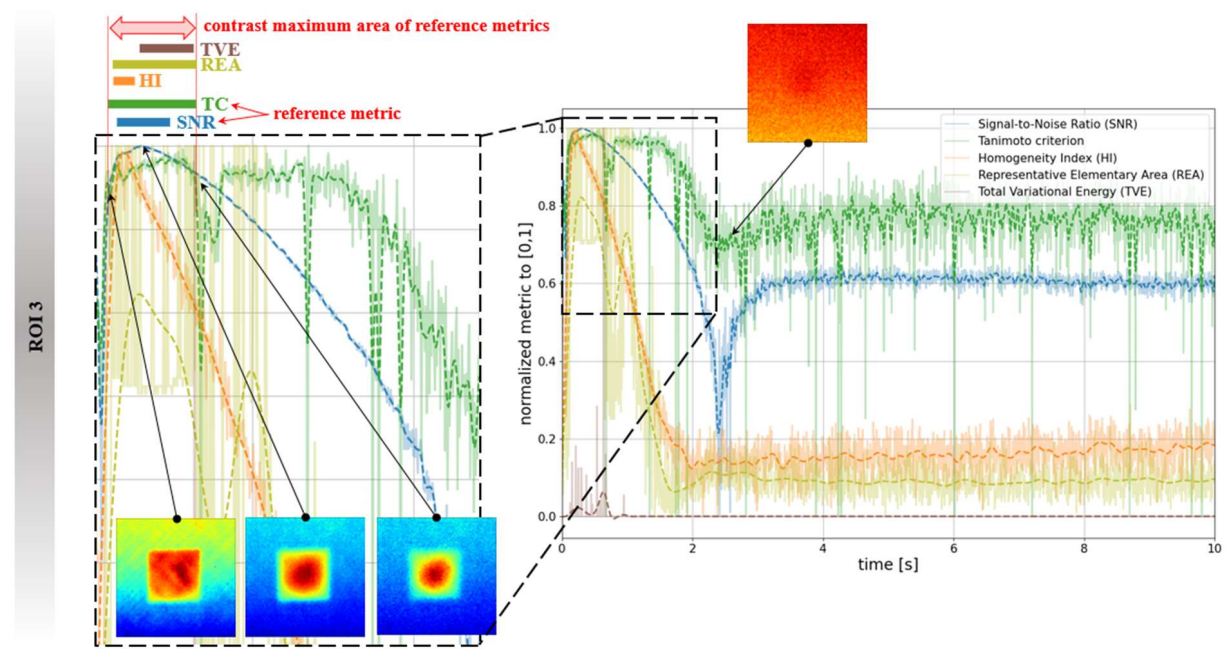

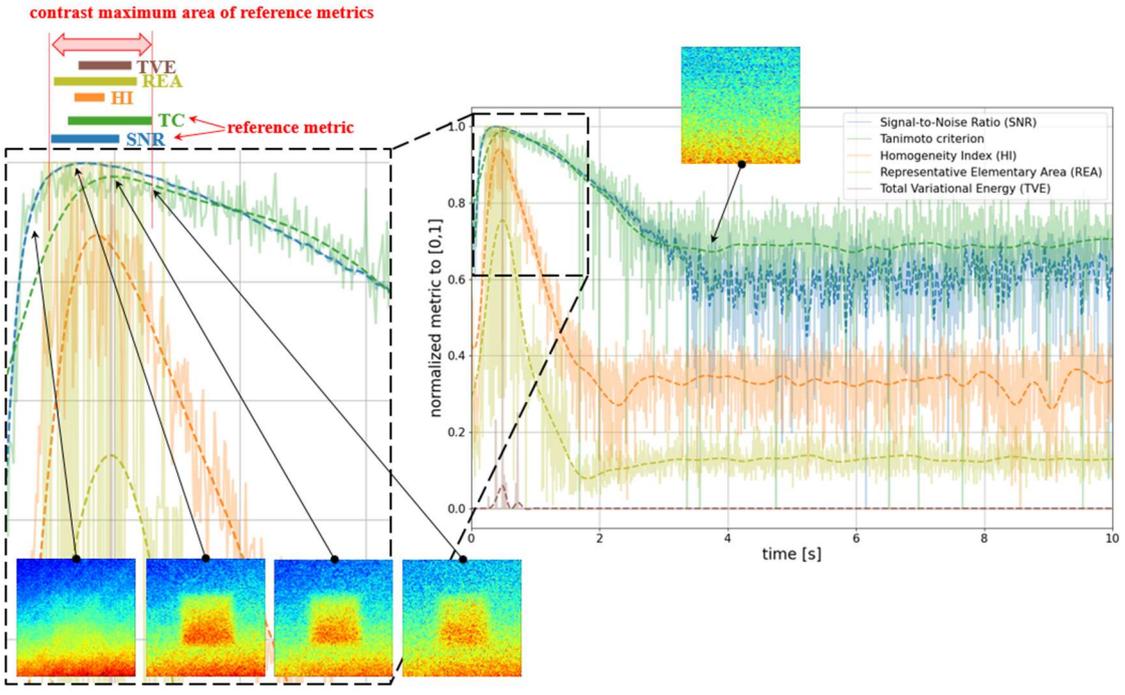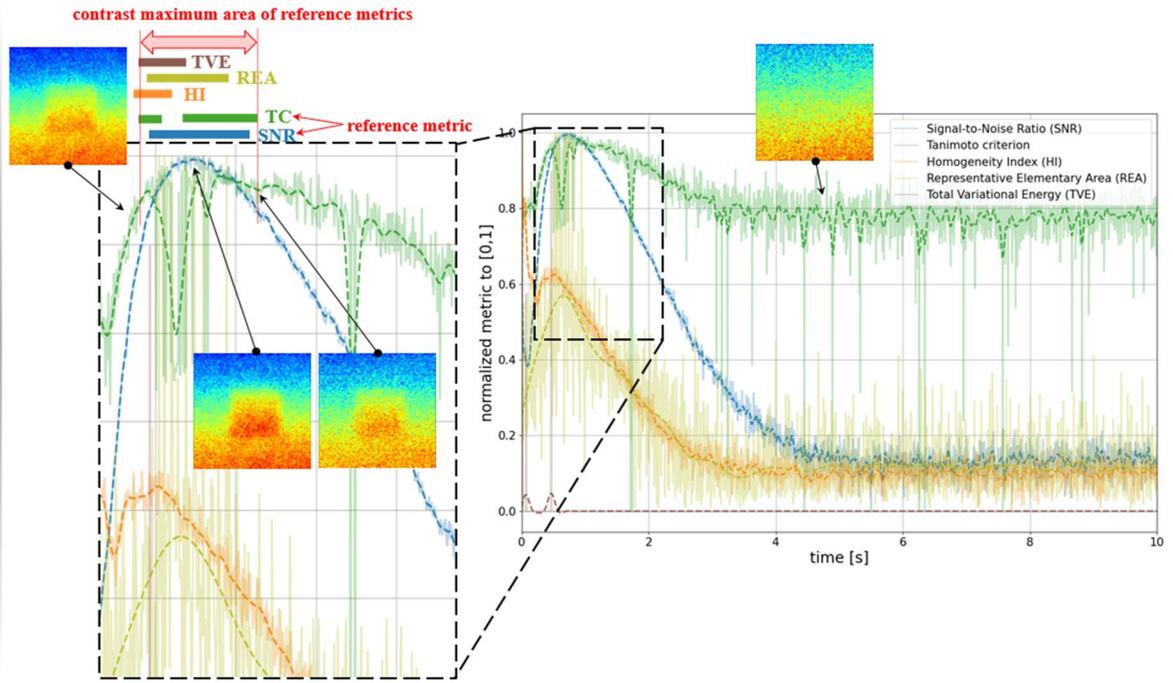

# Appendix B

The subsequent appendix presents the corresponding metric curves for the remaining regions of interest (ROIs) for amplitude and phase sequences, shown analogously to Figure 12. All curves are normalised to the interval [0,1], and identical processing parameters are applied (cluster numbers = 5 for amplitude and 4 for phase sequences, NOS = 439). The filtered metric curves are displayed as dotted lines, while bars indicate index ranges associated with locally and globally maximum metric values.

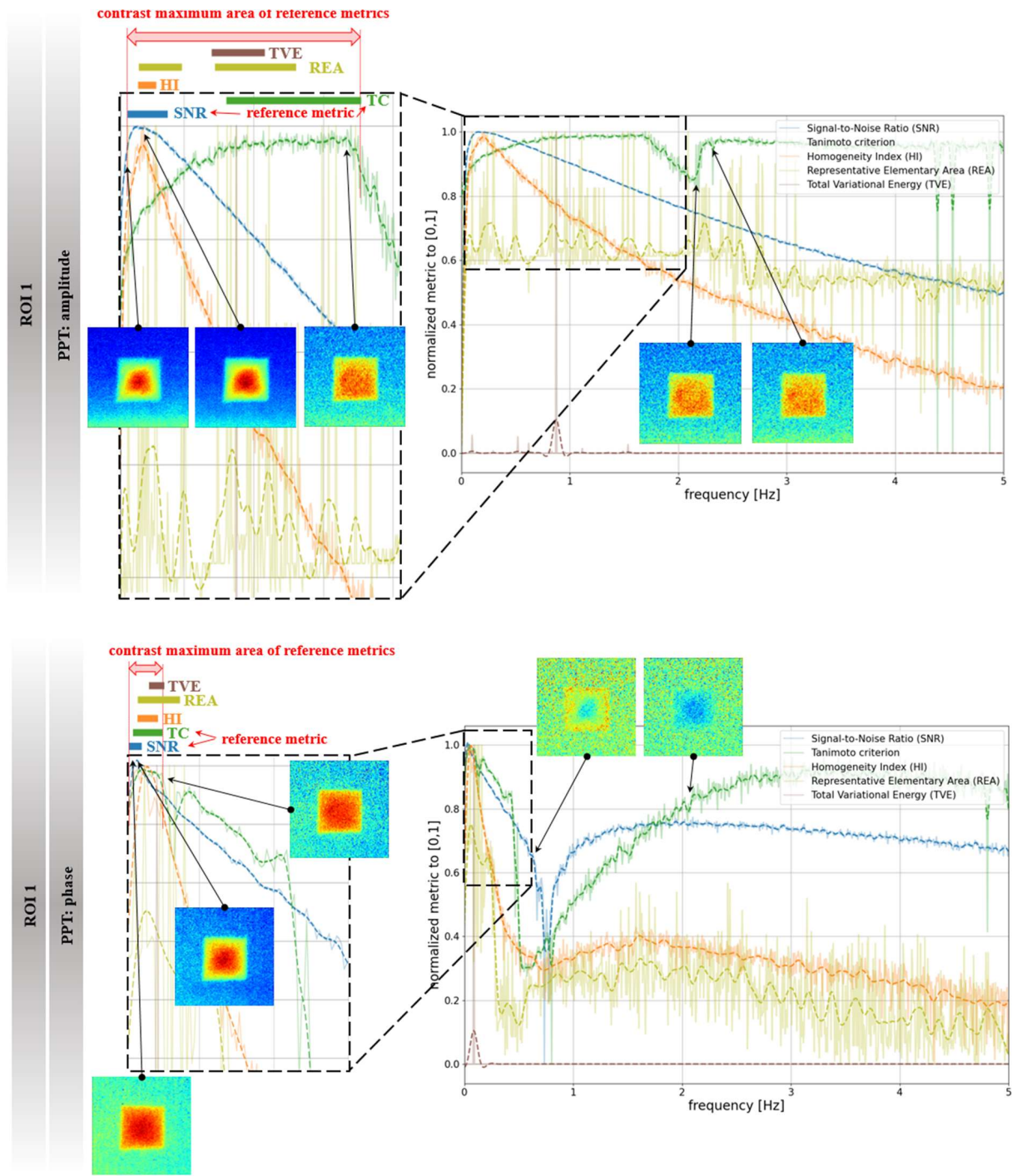

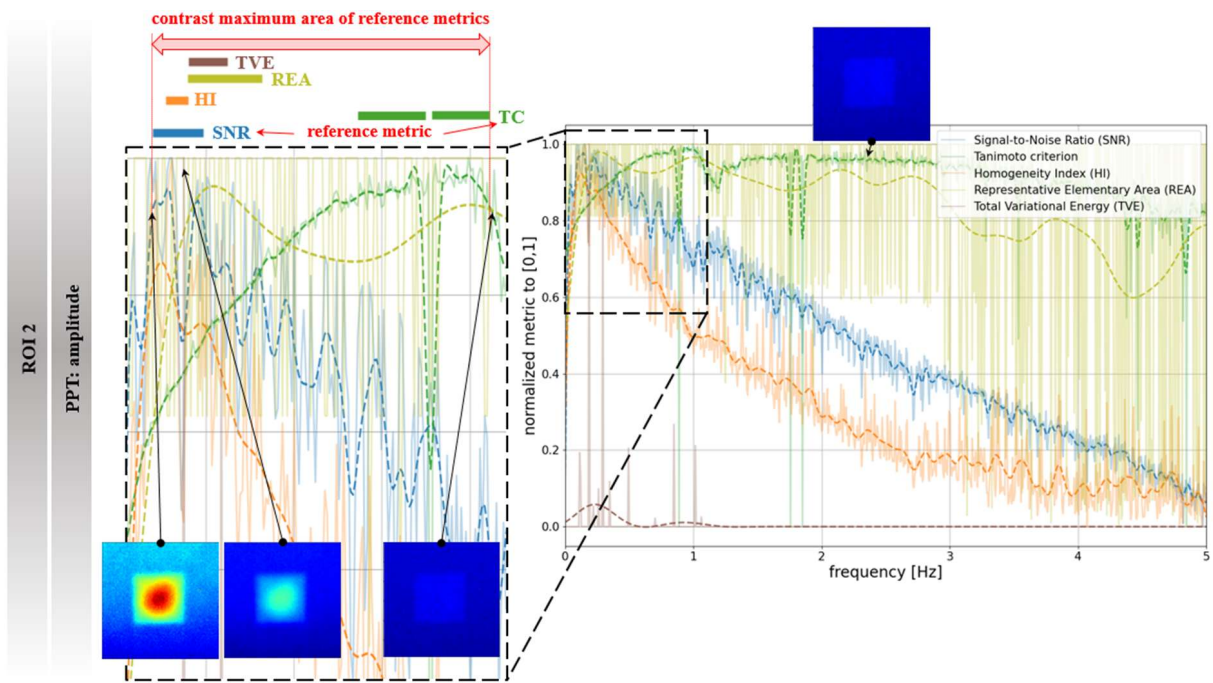

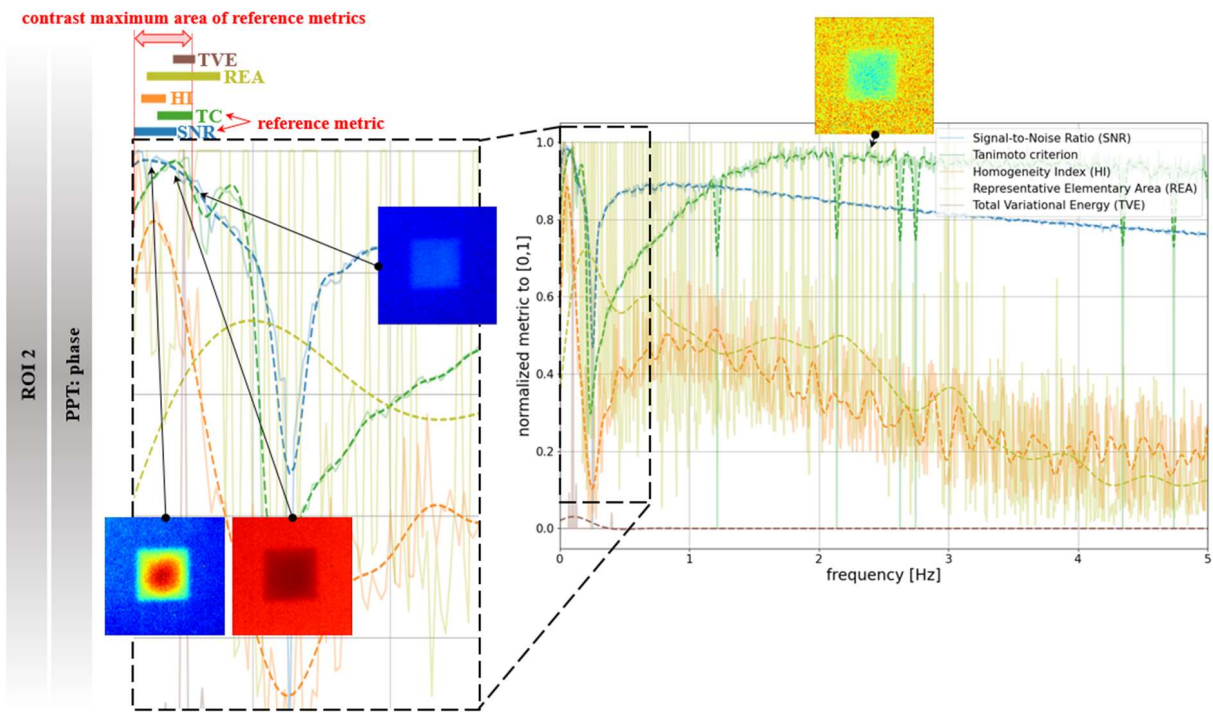

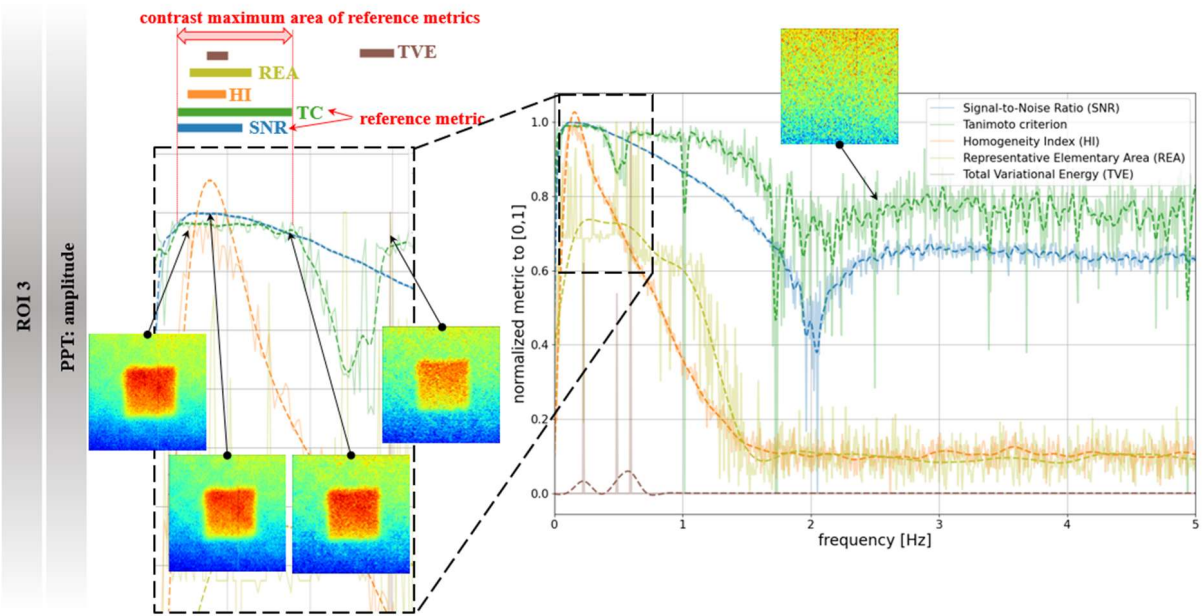
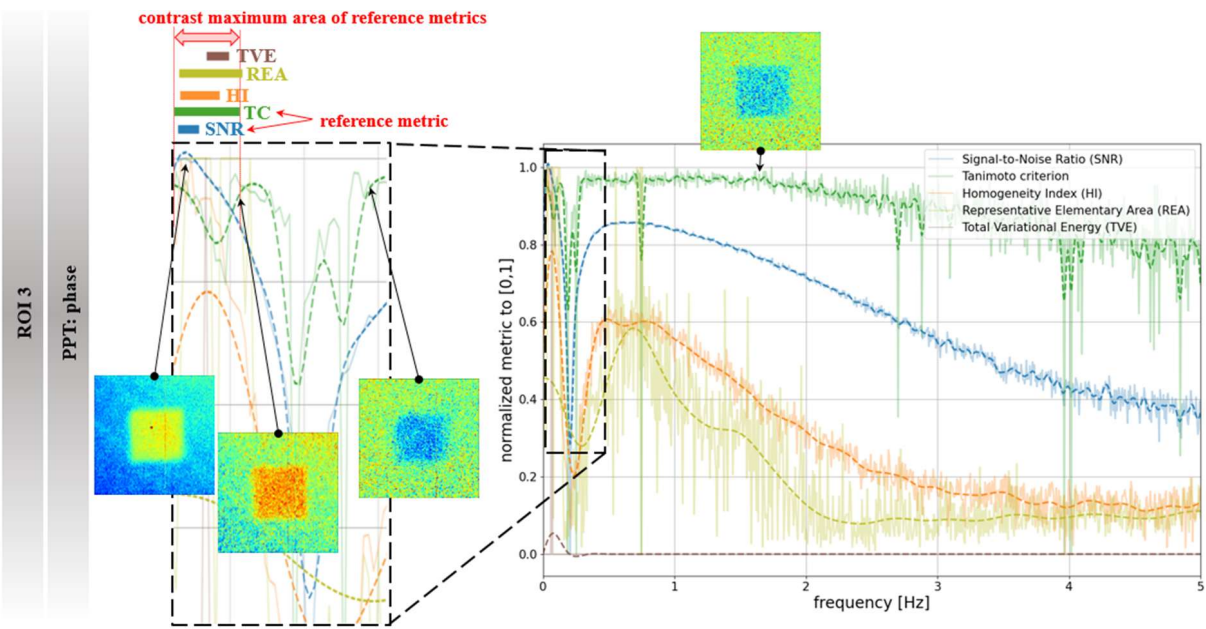

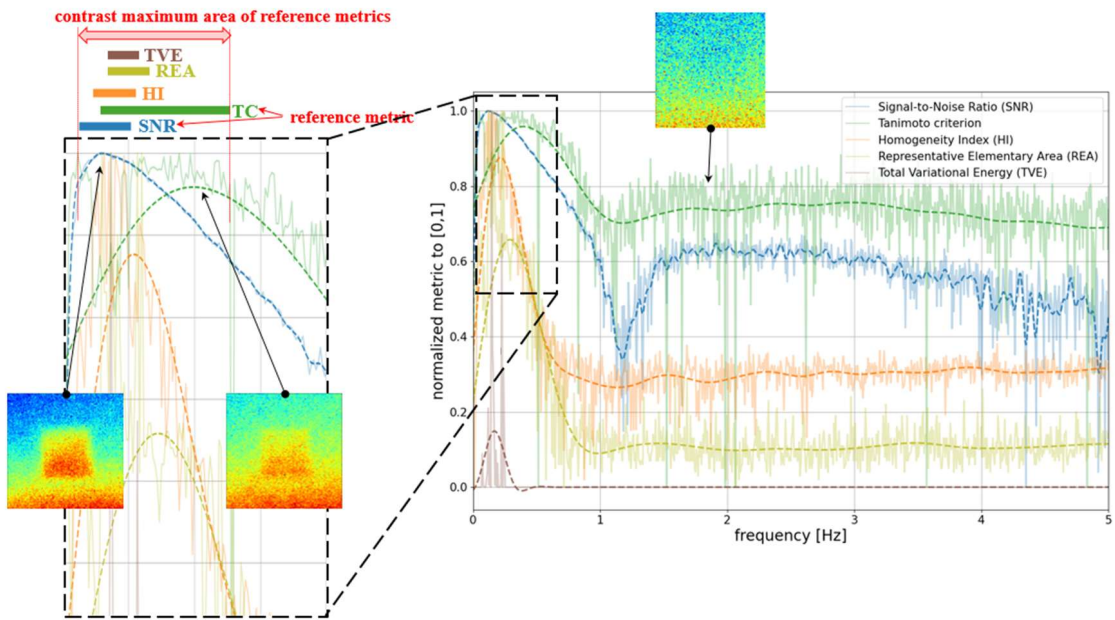
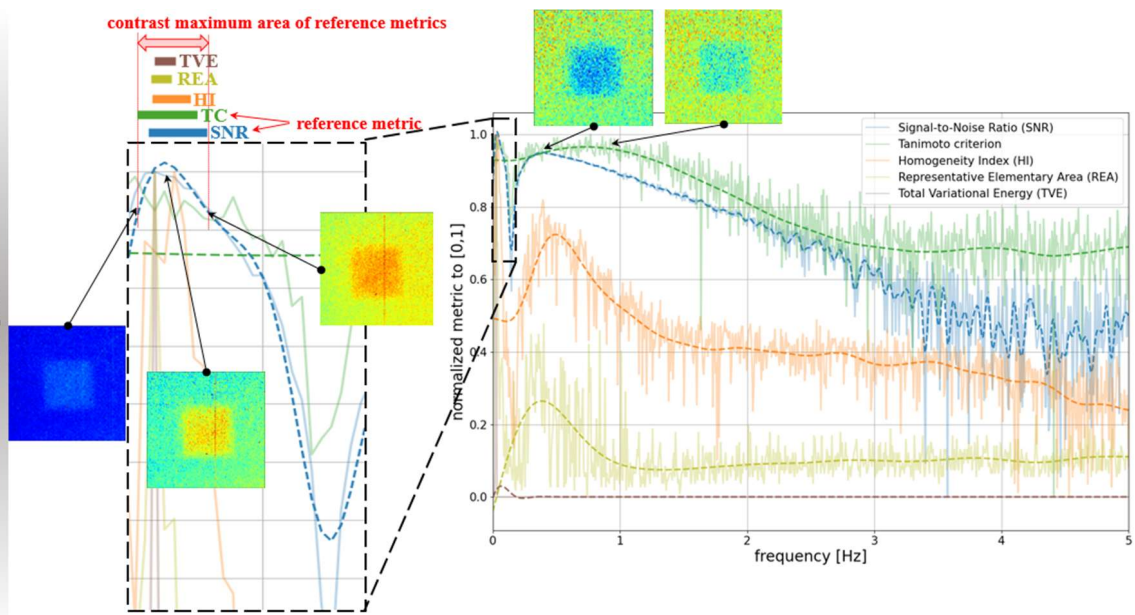

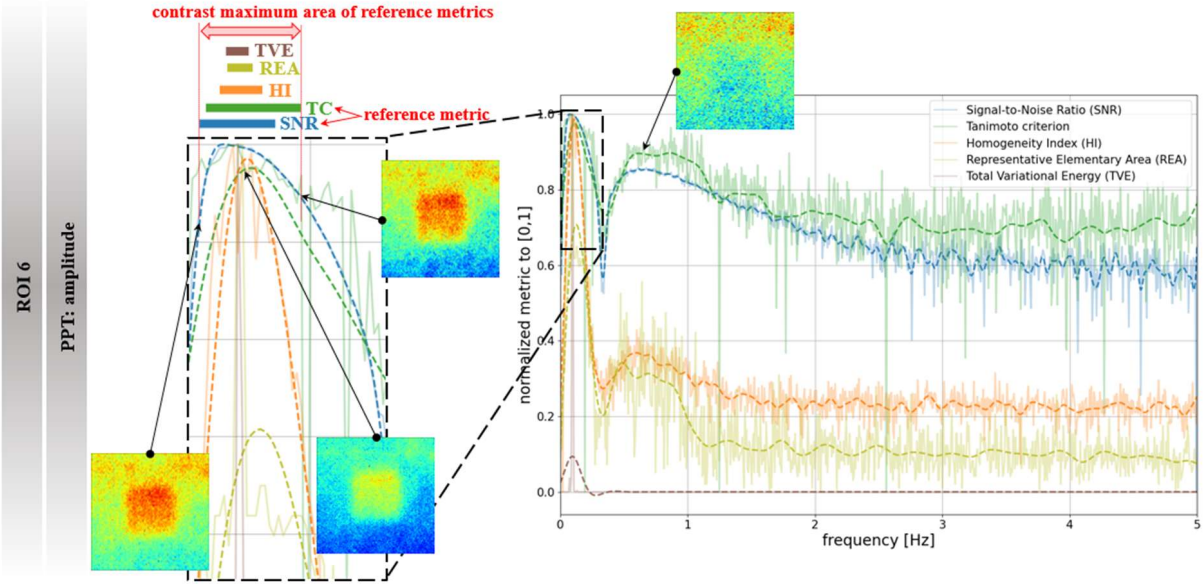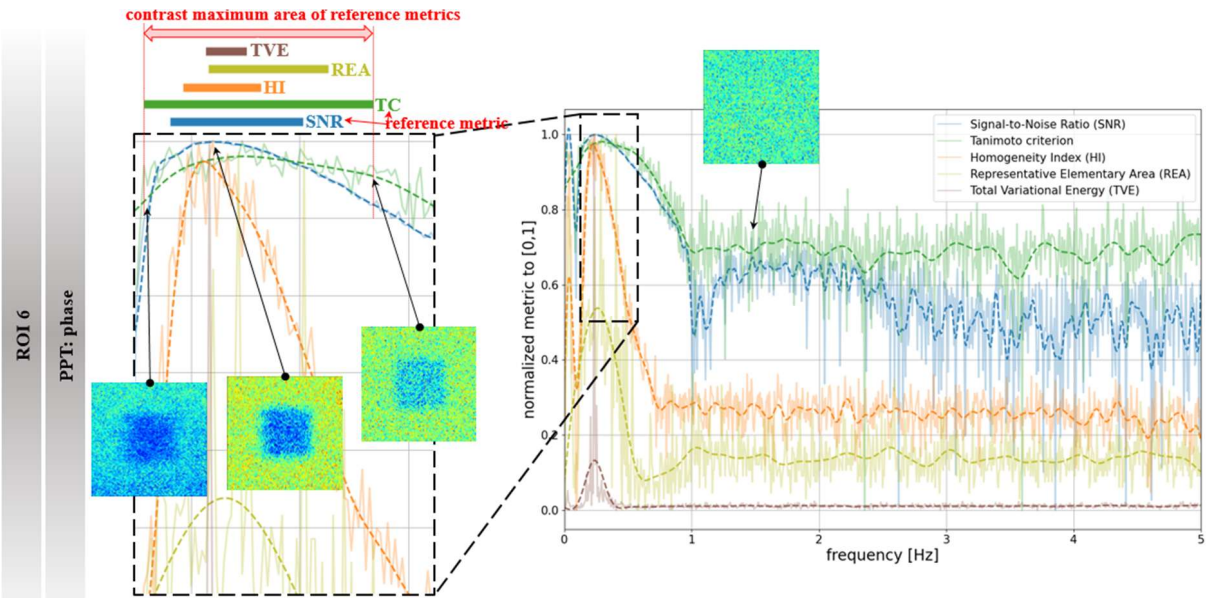

# Appendix C

The present appendix provides a schematic representation of the formation and calculation of the "difference of thermal diffusion length" between the reference and defective areas for the CFRP plate employed in the present study. Firstly, a theoretical thermal simulation is performed for each ROI area (a 1-layer model representing the reference area and a 3-layer model representing the defective area) and the frequency-dependent diffusion depth is calculated. Subsequently, the difference between these values is calculated and displayed in a diagram for all six ROIs. These curves represent the frequency-dependent contrast between the defective and defect-free or reference areas.

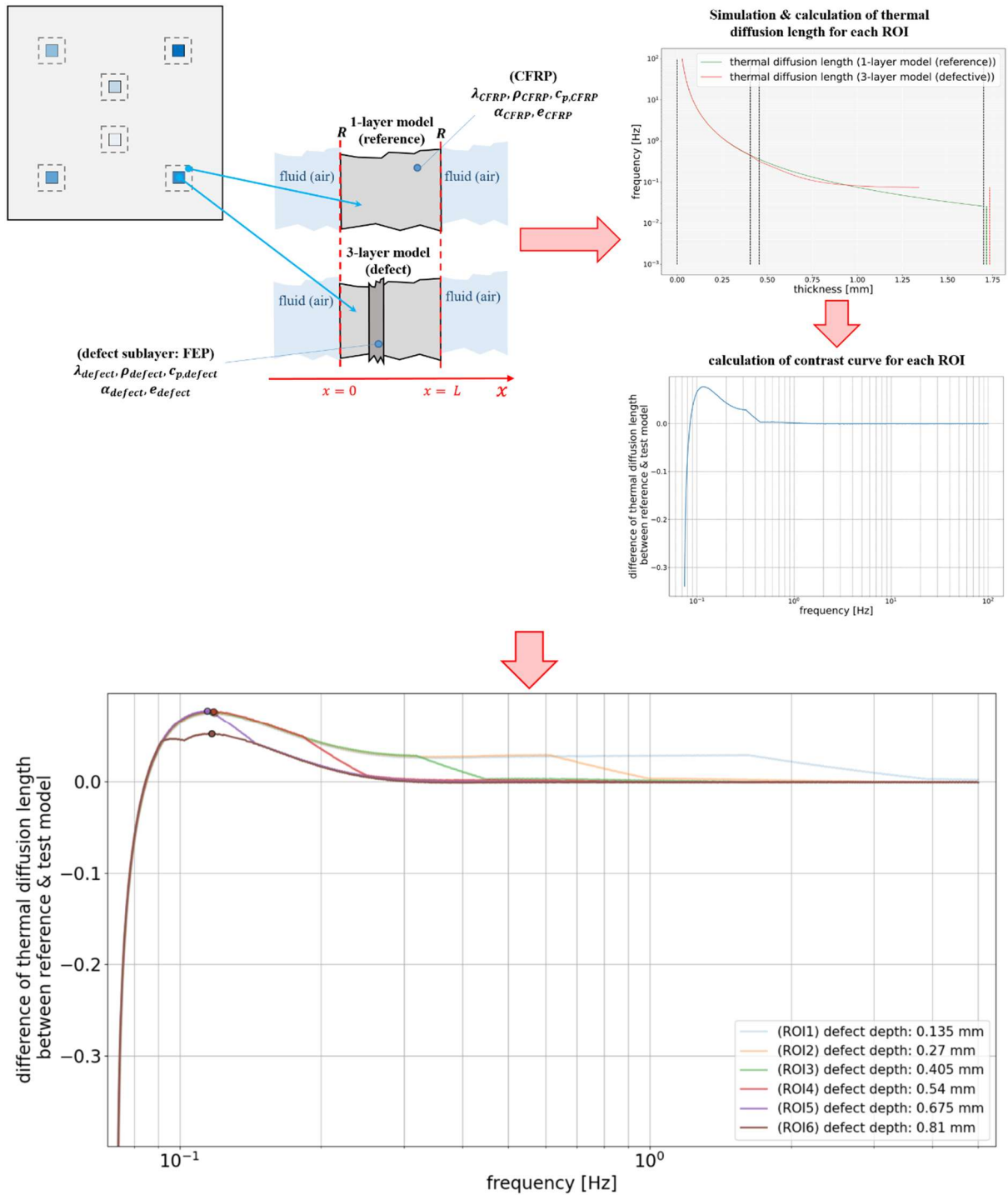

# Appendix D

Python-Version: 3.13.5 | packaged by Anaconda, Inc.

=== Systeminformation ===
System: Windows
Architektur: AMD64
Prozessor: Intel64 Family 6 Model 158 Stepping 10, GenuineIntel
CPU-Kerne (physisch): 6
CPU-Kerne (logisch): 12

=== RAM ===
Total (GB): 63.95
Available (GB): 46.11

=== GPU ===
GeForce GT 1030, 2048 MiB